**Title:**

Integrated Transcriptomic–proteomic Biomarker Identification for Radiation Response Prediction in Non-small Cell Lung Cancer Cell Lines

**Authors:**

Yajun Yu[1,2], Guoping Xu[1,2], Steve Jiang[1,2], Robert Timmerman[1], John Minna[3], Yuanyuan Zhang[1], and Hao Peng[1,2]

**Author affiliations:**

[1]Department of Radiation Oncology, University of Texas Southwestern Medical Center, Dallas, TX 75390, USA.

[2]Medical Artificial Intelligence and Automation Laboratory, University of Texas Southwestern Medical Center, Dallas, TX 75390, USA.

[3]Hamon Center for Therapeutic Oncology Research and the Simmons Comprehensive Cancer Center, University of Texas Southwestern Medical Center, Dallas, TX 75390, USA

*Corresponding author

Email: hao.peng@utsouthwestern.edu

Postal address: Department of Radiation Oncology, University of Texas Southwestern Medical Center, 2280 Inwood Rd., Dallas, TX 75390, USA.

**Abstract**

**Purpose:**

To develop an integrated transcriptome–proteome framework for identifying concurrent biomarkers predictive of radiation response, as measured by survival fraction at 2 Gy (SF2), in non-small cell lung cancer (NSCLC) cell lines.

**Methods and Materials:**

RNA sequencing (RNA-seq) and data-independent acquisition mass spectrometry (DIA-MS) proteomic data were collected from 73 and 46 NSCLC cell lines, respectively. Following preprocessing, 1,605 shared genes were retained for analysis. Feature selection was performed using least absolute shrinkage and selection operator (Lasso) regression with a frequency-based ranking criterion under five-fold cross-validation repeated ten times. Support vector regression (SVR) models were constructed using transcriptome-only, proteome-only, and combined transcriptome–proteome feature sets. Model performance was assessed by the coefficient of determination ($R^2$) and root mean square error (RMSE). Correlation analyses evaluated concordance between RNA and protein expression and the relationships of selected biomarkers with SF2.

**Results:**

RNA–protein expression exhibited significant positive correlations (median Pearson's $r = 0.363$). Independent pipelines identified 20 prioritized gene signatures from transcriptomic, proteomic, and combined datasets, including concurrent biomarkers such as KDM2A, PSIP1, and PTBP2. Models trained on single-omic features achieved limited cross-omic generalizability, while the combined model demonstrated balanced predictive accuracy in both datasets ($R^2 = 0.461$, RMSE = 0.120 for transcriptome; $R^2 = 0.604$, RMSE = 0.111 for proteome). Several biomarkers (e.g., KDM2A, GOT2, CCDC9) exhibited negative correlations with SF2, suggesting associations with radioresistance.

**Conclusions:**

This study presents the first proteotranscriptomic framework for SF2 prediction in NSCLC, highlighting the complementary value of integrating transcriptomic and proteomic data. The identified concurrent biomarkers capture both transcriptional regulation and functional protein activity, offering mechanistic insights and translational potential. The combined model outperformed single-omic approaches, supporting the utility of multi-omic integration for developing robust radiosensitivity predictors and advancing precision radiotherapy.

# 1. Introduction

Lung cancer is the most commonly diagnosed cancer (almost 2.5 million new cases) and the leading cause of cancer death (over 1.8 million) in terms of global cancer statistics in 2022.[1] Among its histological subtypes, non-small cell lung cancer (NSCLC) accounts for approximately 85% of all cases and includes adenocarcinoma, squamous cell carcinoma, and large cell carcinoma.[2-4] Despite therapeutic advances in surgery, systemic therapy, and immunotherapy, radiation therapy (RT) remains central to NSCLC treatment across early-stage, locally advanced, and oligometastatic presentations.[5,6] Clinical outcomes after RT vary dramatically, driven by complex mechanisms of intrinsic or acquired radioresistance involving DNA damage repair, hypoxia responses, oncogenic signaling, and tumor–immune microenvironment interactions.[7-9] As such, identifying robust, predictive biomarkers of RT sensitivity is a critical need for advancing precision radiotherapy in NSCLC.

High-throughput molecular profiling technologies offer promising paths toward biomarker discovery. Transcriptomic analysis, particularly RNA sequencing of protein-coding genes (i.e. messenger RNA or mRNA), has been widely employed to generate gene expression signatures to predict radiation response, as represented by survival fraction at 2 Gy (SF2), in human cancer cell lines.[10-12] However, mRNA expression does not always correlate with functional protein levels due to post-transcriptional and post-translational regulation, limiting interpretability from transcriptome data alone.[13] In contrast, proteomic profiling offers direct insight into signaling cascades, DNA repair machinery, and stress response pathways that determine cellular radiosensitivity.[14,15] Nevertheless, proteomic measurements alone lack the breadth and scalability of transcriptomic surveys. Although identification of proteomic biomarkers has been significantly explored on radiation-induced changes in protein abundance associated with biological functions, such as the DNA damage response and stress response,[16-18] no proteome-related prediction model, to the best of our knowledge, has been reported to estimate radiation response across pan-cancer or tissue-specific cell lines.

Integrating both the protein-coding transcriptome and proteome in the same biological samples is emerging as a powerful strategy to identify concurrent biomarkers that reflect both transcriptional regulation and functional protein activity. In NSCLC cell lines, plenty of proteogenomic studies have demonstrated how combining RNA-sequencing and mass-spectrometry proteomics uncovers molecular subtypes, and enables discovery of protein-level aberrations not fully captured by transcript data, thereby enhancing biological interpretability and predictive potential.[19,20] For example, comparative proteotranscriptomic profiling has identified proteins whose expression is tightly correlated with their corresponding mRNAs, such as ANXA1, ANXA2, and HNRNPA2B1, highlighting markers with concordant RNA and protein levels that may serve as more robust biomarkers.[21]

In this work, a transcriptome-proteome combined model was developed to identify concurrent protein biomarkers for radiation response prediction in 73 NSCLC cell lines. By leveraging a dual-omics framework, the selected concurrent genes were employed to establish support vector regression (SVR) models for SF2 prediction in both transcriptomic and proteomic datasets. Compared with individual-omics approaches, which ruined the prediction accuracy in the other omic dataset, the combined model guaranteed the predictive powers simultaneously, underscoring the value of concurrent biomarker discovery in capturing complementary layers of molecular information. This study firstly presented a proteotranscriptomic model for SF2 estimation across NSCLC cell lines, providing a foundation for future pan-cancer or other specific-cancer application and clinical translation.

## 2. Methods and Materials

### 2.1 Transcriptome and proteome datasets

The transcriptome expression profiles were RNA sequencing (RNA-seq) from 73 NSCLC cell lines. A total of 19,411 protein-coding gene signatures were extracted for each NSCLC cell line. The proteome dataset was obtained from an online available resource-ProCan-DepMapSanger, including 948 human pan-cancer cell lines analyzed by data independent acquisition-mass spectrometry (DIA-MS), resulting in 8,453 gene signatures.[22]

### 2.2 Clonogenic cell survival assays

Exponentially growing NSCLC cells were harvested by trypsinization, resuspended in growth medium, and adjusted to obtain single-cell suspensions. Cell numbers were determined, and an appropriate number of cells was plated for each radiation dose (with higher doses requiring more cells). Cells were seeded in 60-mm tissue culture dishes in triplicate for each dose and allowed to attach to the dish by incubating them for 6-8 hours before irradiation. Irradiation was performed at the indicated doses (0, 1, 2, 3, and 4 Gy) using a 137Cs Mark 1-68 irradiator (J.L. Shepherd and Associates, San Fernando, CA). Following treatment, plates were incubated for 7–14 days to allow colony formation. Colonies were fixed with 4% formaldehyde in PBS and stained with 0.05% crystal violet. Colonies containing more than 50 cells were counted manually under a microscope. The surviving fraction was calculated as:

$$\text{Survival fraction} = \frac{\text{(Mean colonies counted)}}{\text{(Cells plated)} \times \text{(Plating efficiency)}}$$

where plating efficiency was defined as

$$\text{Plating efficiency} = \frac{(\text{Mean colonies connted})_{\text{control}}}{(\text{Cells plated})_{\text{control}}}$$

where "control" denotes unirradiated control for cells. All experiments were repeated three times, and data were reported as averaged values. Survival curves were generated by curve fitting to the standard linear–quadratic model using SigmaPlot (Systat Software Inc.). SF2 values were obtained from fitted curves.

**2.3 NSCLC Cell line selection**

Cell line selection was based on the match of transcriptome or proteome dataset with existing 73 SF2 values of NSCLC cell lines (see Cell Line Selection module in **Figure 1**). The whole 73 cell lines from transcriptome dataset coincide with SF2 values, while a subgroup of proteomic pan-cancer cell lines match with the measured radiation responses, bringing forth 46 mutual cell lines. For the correlation analysis between RNA-seq and proteomic expression, transcriptomic dataset was restricted to the same 46 cell lines to ensure sample consistency with the proteomic cohort.

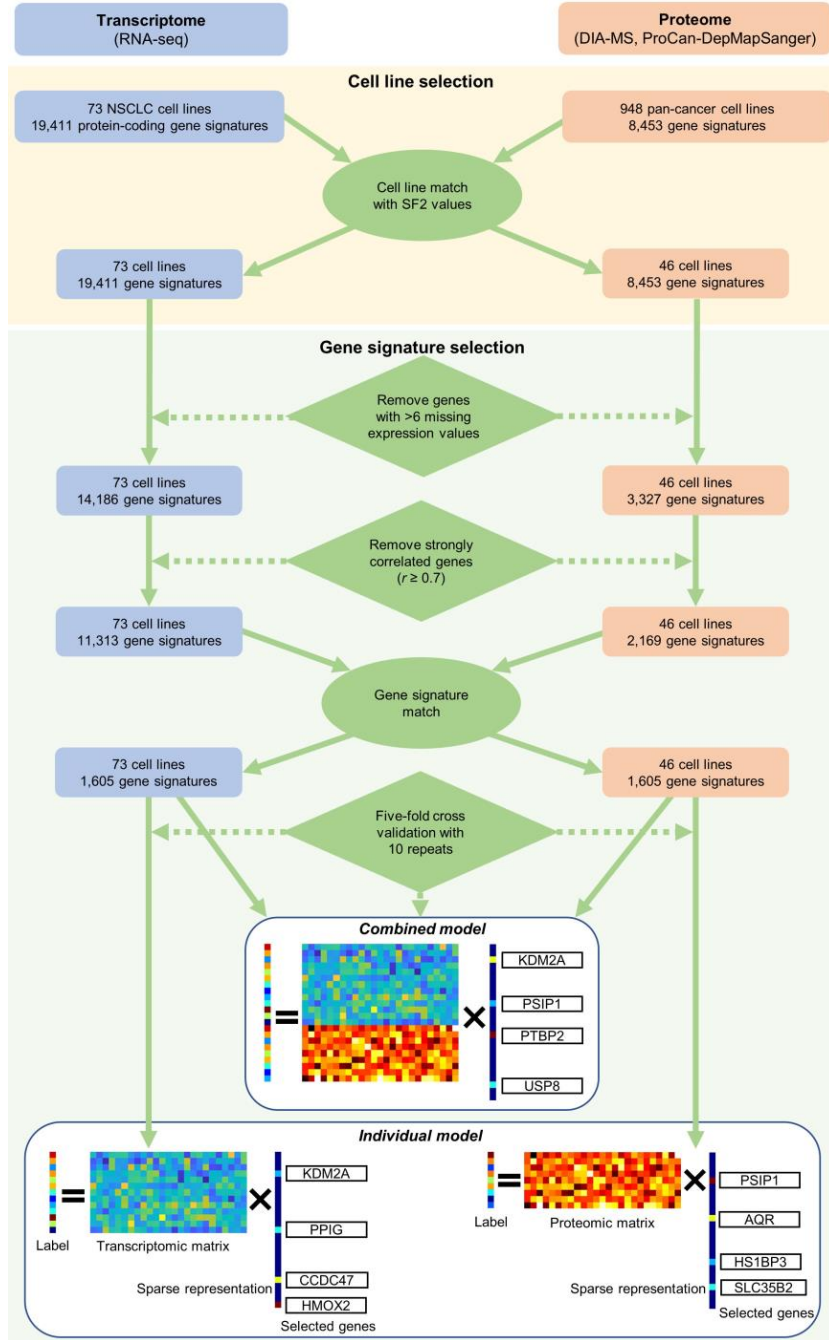

**Figure 1.** Flowchart of cell line selection and gene signature selection for transcriptome and proteome datasets.

**2.4 Gene signature selection**

Two separate gene selection pipelines were implemented for transcriptome and proteome datasets (see Gene Signature Selection module in **Figure 1**). A mixture of filtering and feature-reduction strategies was applied. First, low-confidence features were excluded by discarding any gene vector with more than six missing expression values across samples. The remaining missing values were replaced by the mean expression of available values in their corresponding gene vector. The remaining missing values in one gene vector was replaced by the mean expression of non-missing values. Second, Pearson linear correlation analysis was performed to remove any redundant gene with strongly absolute correlation coefficient (≥0.7) to any of its previous genes. Third, overlapping features between transcriptomic and proteomic sets were extracted, yielding 1,605 shared gene signatures retained for subsequent modeling. To make all gene vectors comparable, each gene vector was Z-score normalized. Fourth, the label (SF2) vector can be represented by a feature matrix multiplied by a sparse vector with a small number of non-zero coefficients. A feature selection approach using least absolute shrinkage and selection operator (Lasso) algorithm with frequency-based ranking criterion was employed, as described in our previous work.[23] To mitigate the potential overfitting and bias due to the small sample size, a five-fold cross-validation procedure with 10 repeats was executed for further feature selection. In each iteration, a fixed number of 30 non-zero coefficients were determined by tuning the regularization parameter in Lasso, leading to a reservation of 30 corresponding feature candidates. The top 20 features with maximum absolute coefficients from 30 candidates were counted once per iteration, followed by frequency-based ranking across whole iterations. 20 genes were ultimately selected and ranking according to their importance scores averaged through 50 iterations. Lastly, two scenarios of feature sets were evaluated in this study: (i) individual model, using only transcriptomic or proteomic feature matrix, and (ii) combined model, constructed by merging transcriptomic and proteomic feature matrices along the shared gene signature dimension.

**2.5 SF2 prediction model**

Support vector regression (SVR), an extension of the support vector machine (SVM) framework, was employed to model SF2 values. SVR maps input features into a high-dimensional space to construct a continuous-valued function within an ε-insensitive margin, penalizing training samples that fall outside this tolerance region.[24, 25] The transformation of feature spaces was achieved using kernel functions. Linear kernel function was applied in this study. The best scaling parameter in each SVR model was determined by the coefficient of determination ($R^2$) through the grid search with 3-fold cross-validation. The scaling parameter grid was in a range of $[2^{-10}, 2^{-9}, 2^{-8}, \ldots, 2^{10}]$. The other parameters are set as default in the sklearn.svm.SVR.[26]

### 2.6. Statistical analysis

Regression model performance was assessed using two metrics: the coefficient of determination ($R^2$) and the root mean square error (RMSE). The definitions are as follows:

$$R^2 = 1 - \frac{\sum_{i=1}^{n}(y_i - \hat{y}_i)^2}{\sum_{i=1}^{n}(y_i - \bar{y})^2}$$

$$RMSE = \sqrt{\frac{\sum_{i=1}^{n}(y_i - \hat{y}_i)^2}{n}}$$

where $y_i$ is the measured SF2 value of the *i*-th sample, $\hat{y}_i$ is the predicted SF2 value of the *i*-th sample, $\bar{y}$ is the mean of all measured SF2 values, and *n* is the number of samples. $R^2$ quantifies the proportion of variance in the measured SF2 data explained by the model, with values closer to 1 indicating stronger predictive performance. RMSE measures the average magnitude of prediction errors, with lower values reflecting higher predictive accuracy.

To minimize bias in prediction and to obtain empirical estimates of performance variability, a five-fold cross validation with 10 repeats was employed in the regression model construction. This procedure generated 50 independent estimates for each performance metric. Final model performance was reported as the mean value across repetitions, accompanied by 95% confidence intervals (CIs). Pearson linear correlation coefficient between two variables was calculated to evaluate the degree of their relationship. Variance inflation factor (VIF) was used to examine the multicollinearity issue in a regression model.[27] All analyses were implemented using MATLAB R2023b and Python v3.9.18.

### 3. Results

#### 3.1 Correlation between RNA-seq and proteomic expressions

To investigate the concordance between transcriptomic and proteomic data by individual genes, RNA-seq expression levels were compared with mass-spectrometry-based proteomic abundances for matched 1,605 genes across 46 NSCLC cell lines. As shown in **Figure S1** (Supporting Information), most protein-coding genes demonstrated significant positive correlations between RNA and protein levels, with a median Pearson's *r* = 0.363. The high consistency of RNA-seq and proteomic expression underscores the importance of concurrent biomarker identification for radiation response prediction, enabling predictive signatures that capture both transcriptional regulation and functional protein activity.

## 3.2 Selected protein biomarkers

Feature selection pipelines were applied independently to the protein-coding transcriptomic, proteomic, and combined dataset, yielding three sets of 20 prioritized gene signatures (**Table 1**). These included both transcriptome-derived (e.g., *KDM2A*, *PPIG*, *CCDC47*, *HMOX2*) and proteome-derived (e.g., *PSIP1*, *AQR*, *HS1BP3*, *SLC35B2*) genes, as well as the concurrent protein biomarkers from the combined model (e.g., *KDM2A*, *PSIP1*, *PTBP2*, *USP8*). Comparing with two series of selected gene signatures from individual models, merely *KDM2A* was repeated, ranking 1st in the transcriptomic panel and 11th in the proteomic panel, while the remaining genes were unique to their respective omics layers. Within the combined model, several genes were overlapped in the transcriptomic panel (i.e., *KDM2A*, *GOT2*, *TRMT1L*, *SMC6*, and *CCDC9*) and proteomic panel (i.e., *KDM2A*, *PSIP1*, *PTBP2*, *TRNIIP1*, *NCOR2*, and *MRPS14*), emphasizing the complementary nature of transcriptome-proteome combined data in capturing radiosensitivity-associated biomarkers.

**Table 1** Top 20 gene signatures selected from transcriptome-only, proteome-only, and transcriptome-proteome combined feature selection pipelines. Rankings reflected importance scores averaged across 50 iterations. The gene substitutes were included in brackets.

| Importance ranking | Individual model | | Combined model |
|---|---|---|---|
| | Only transcriptome-selected genes | Only proteome-selected genes | |
| 1 | KDM2A | PSIP1 | KDM2A |
| 2 | PPIG | AQR | PSIP1 |
| 3 | CCDC47 | HS1BP3 | PTBP2 |
| 4 | HMOX2 | SLC35B2 | USP8 |
| 5 | POLR2E | GOSR1 | GOT2 |
| 6 | TMEM165 | MRPS14 | TECPR2 |
| 7 | ENY2 | CD276 | RTN4IP1 |
| 8 | BPHL | PRPF40A | TRMT1L (ACOT13) |
| 9 | HMGCL | GOLPH3 | NCOR2 |
| 10 | GOT2 | DHX8 | CCDC9 |
| 11 | TRMT1L | KDM2A | SMC6 |
| 12 | SMC6 | KIDINS220 | MRPS14 (NDUFB1) |
| 13 | CLASP1 | RTN4IP1 | TMEM214 |
| 14 | PLGRKT | BCS1L | CLTC |
| 15 | MTDH | CCAR1 | TST (MPST, TKT) |
| 16 | HINT2 | NCOR2 | LACTB2 |
| 17 | CDC40 | GNL1 | MLEC |
| 18 | CCDC9 | POLR2F | RAB3GAP2 |
| 19 | EPPK1 | PTBP2 | AGRN |

| 20 | TMED1 | WDR75 | HSPG2 |

Given the complex and interconnected nature of gene regulation, excluded genes may still possess predictive potential. To account for this, substitution candidates were identified based on strong correlations with both the corresponding gene vector ($r \geq 0.4$) and the SF2 label vector ($r \geq 0.2$). Therefore, in the combined model, *TRMT1L* was able to be substituted with *ACOT13*, *MRPS14* with *NDUFB1*, and *TST* with *MPST* or *TKT*.

**3.3 SF2 regression model performance**

SVR models for SF2 prediction were developed using the transcriptomic, proteomic, and combined gene sets, and their predictive performances were evaluated by $R^2$ and RMSE under five-fold cross-validation with 10 repeats, as illustrated in **Figure 2**. Regression models trained on only transcriptome- or proteome-selected features performed reasonably well within their own datasets but failed to generalize across the other omic layer. For instance, only transcriptome-trained models had an $R^2 = 0.572$ using top 18 selected genes for transcriptomic data, whereas the $R^2$ dropped to nearly zero when applied to proteomic data. In contrast, the combined model demonstrated robust cross-omic performance, reaching its best accuracy at 14 selected genes in both transcriptome ($R^2 = 0.461$, RMSE = 0.120) and proteome ($R^2 = 0.604$, RMSE = 0.111) datasets. Notably, replacing several combined-model features with their substitution candidates (*TRMT1L* substituted with *ACOT13*, *MRPS14* with *NDUFB1*, and *TST* with *MPST*) yielded prediction performances comparable to the original combined model. Additionally, **Figure 3** exhibited the measured SF2 values and the predicted values derived from the combined model using top 14 selected genes, meanwhile these SF2 values were summarized in **Table S1** (Supporting Information).

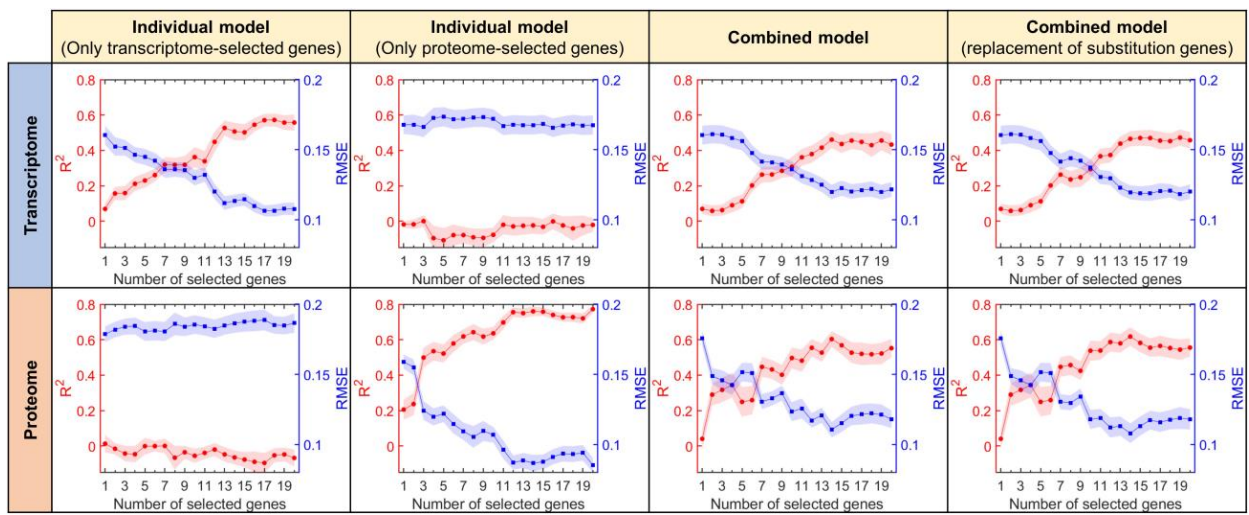

**Figure 2.** Regression evaluation metrics ($R^2$ and RMSE) for individual models (using only transcriptome- or proteome-selected genes) and combined models (using transcriptome-proteome combined selected genes or their substitution candidates) as a function of the number of selected genes. The shaded areas represented 95% CIs.

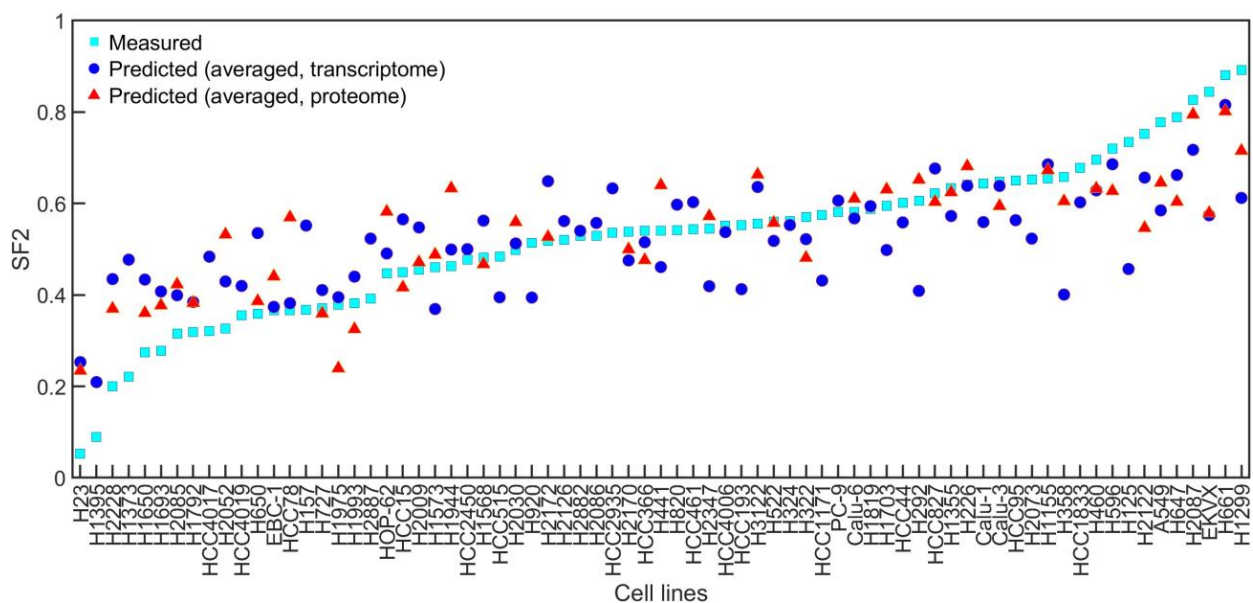

**Figure 3.** Measured SF2 values and predicted values for transcriptome (73 cell lines) and proteome (46 cell lines) datasets by the combined model using top 14 selected genes.

### 3.4 Biomarker correlation analysis

The linear correlation coefficients between the expression levels of the 20 selected genes (derived from the combined model) and the SF2 label vector for transcriptome and proteome datasets were shown in **Figures 4A** and **4B**, respectively. In both datasets, six concurrent biomarkers (i.e., ranking #1: *KDM2A*, #5: *GOT2*, #10: *CCDC9*, #14: *CLTC*, #17: *MLEC*, and #18: *RAB3GAP2*) exhibited negative correlations with SF2, suggesting their higher expressions associated with increased radioresistance. Conversely, the remaining concurrent biomarkers displayed positive associations with SF2, reflecting potential radiosensitive phenotypes. Furthermore, the averaged feature weights assigned by SVR models (**Figures 4C** and **4D**) aligned well with the positive/negative direction of their correlations relative to SF2 (**Figures 4A** and **4B**), except for #15 gene in transcriptome dataset. Notably, the selected biomarkers did not rank consistently with those based on correlation coefficients to the label vector or SVR feature weights, underscoring that the biomarker prioritization based on the feature selection pipeline was influenced by multi-gen interactions rather than individual gene effects alone.

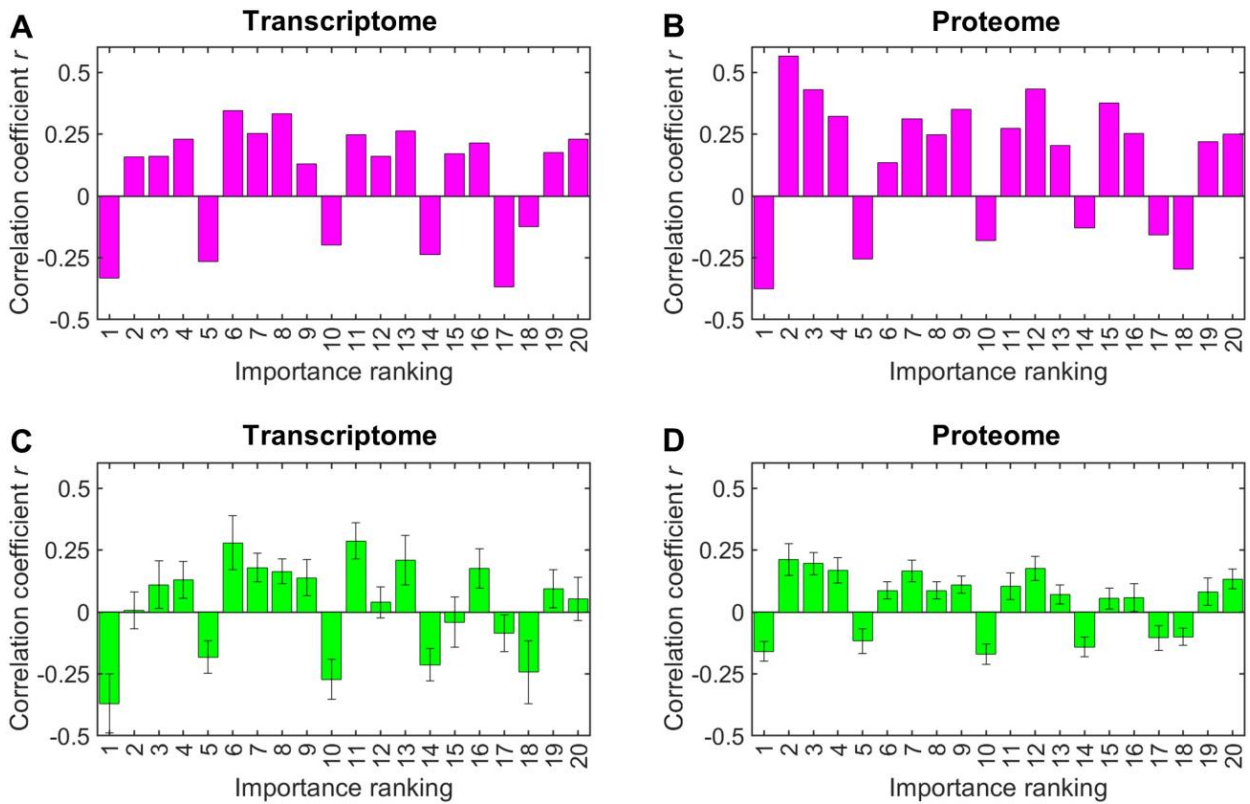

**Figure 4.** Linear correlation coefficients between the expression levels of the 20 selected genes (derived from the combined model) and the SF2 label vector for (**A**) transcriptome and (**B**) proteome datasets. Averaged feature weights assigned by SVR models across 50 iterations for (**C**) transcriptome and (**D**) proteome datasets. Error bars represented 95% CIs.

**Figure 5** exhibited the pairwise correlation heatmap of the top 20 features (derived from the combined model) and the distribution of correlation coefficients (inset) for each dataset, as well as the average of absolute coefficients accompanied with their standard deviation (SD). For the transcriptomic data, the overall pairwise feature association ($\overline{|r_i|} = 0.148$) was slightly higher compared with that of proteomic data ($\overline{|r_i|} = 0.139$). Based on the SDs (transcriptome: $\sigma = 0.112$; proteome: $\sigma = 0.101$) of absolute coefficients and their histogram distributions, transcriptomic coefficients were in a range of [-0.5, 0.6], more diverse than those from the proteome (in [-0.4, 0.4]). Moreover, multicollinearity analysis for the selected feature set and its corresponding set replaced by substitution candidates was assessed using VIF, as shown in **Figures S2** and **S3** (Supporting Information). All VIF values were below 5, indicating weak multicollinearity existing among each feature set.

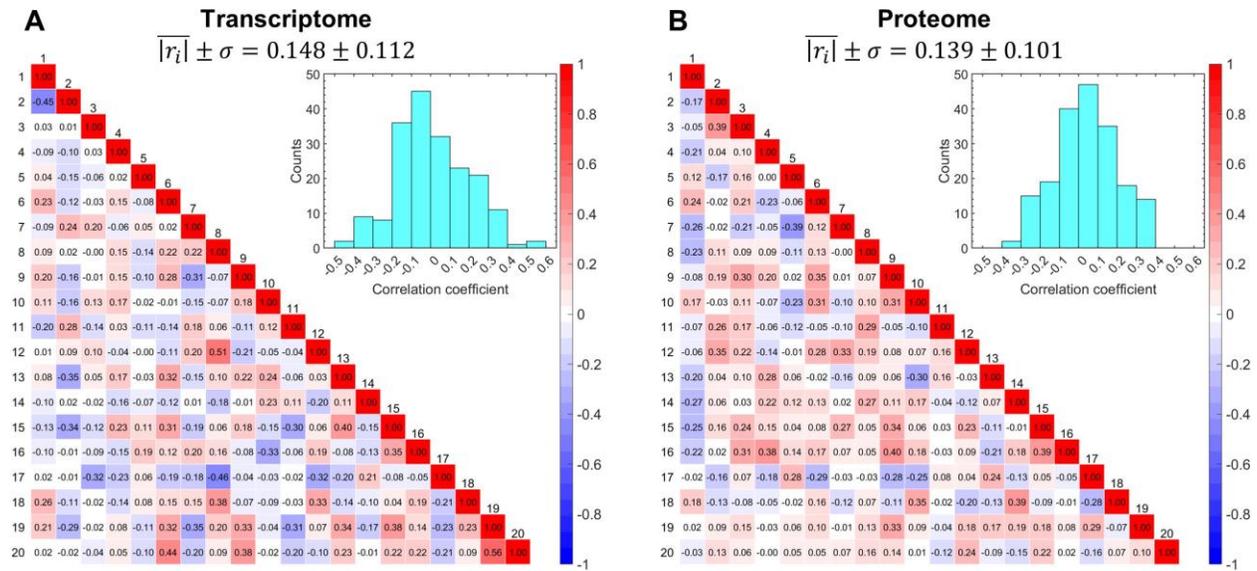

**Figure 5.** Pairwise correlation heatmap of the top 20 features and the average of absolute coefficients accompanied with their standard deviation for transcriptome (**A**) and proteome (**B**) datasets. The distribution of correlation coefficients was displayed in the inset.

## 4. Discussion

In this study, we developed and validated a transcriptome–proteome combined framework to identify concurrent protein biomarkers and predict radiation response, as measured by SF2, in NSCLC cell lines. To our knowledge, this is the first report of a proteotranscriptomic prediction model for radiosensitivity in lung cancer, highlighting the complementary strengths of transcriptomic and proteomic profiling in biomarker discovery.

The correlation analysis for the concordance between RNA-seq and proteomic expressions confirmed that RNA and protein levels exhibit modest but significant concordance (median Pearson's $r = 0.363$), consistent with previous breast-cancer and pan-cancer studies that report similar correlation coefficients (~0.35-0.38).[28, 29] This observation reinforces the need for integrative analyses, since reliance on transcriptomics alone may overlook post-transcriptional regulation, alternative splicing, or protein degradation processes that critically shape radiation response. The biomarker correlation analyses provided further biological insight. A subset of concurrent biomarkers (e.g., *KDM2A*, *GOT2*, *CCDC9*, *CLTC*, *MLEC*, and *RAB3GAP2*) displayed negative associations with SF2, suggesting that their elevated expression contributes to radioresistance. Conversely, other candidates were positively correlated with SF2, indicative of radiosensitizing roles. These bidirectional patterns highlight the complex interplay of gene networks in shaping radiation response, where both DNA repair promotion and stress pathway suppression can influence outcome.

Through independent and joint feature selection pipelines, we identified reproducible sets of transcriptome-derived, proteome-derived, and concurrent biomarkers. Interestingly, only one gene (*KDM2A*) was repeatedly selected across both individual omics models, underscoring the unique and complementary contributions of each molecular layer. Within the combined model, shared biomarkers such as *KDM2A*, *PSIP1*, and *PTBP2* emerged as strong candidates for radiosensitivity regulation. *KDM2A*, for example, is a histone demethylase implicated in chromatin remodeling and DNA damage repair, processes closely tied to radiation response. Similarly, *PSIP1* (LEDGF/p75) is known to mediate stress response signaling and transcriptional co-activation, while *PTBP2* regulates RNA splicing events that may influence DNA repair pathways. Predictive modeling further highlighted the advantages of integrative analysis. SVR models trained exclusively on transcriptomic or proteomic features achieved reasonable performance in their own datasets but failed to generalize across the other omic layer, suggesting limited transferability of single-omics predictors. In contrast, the combined model demonstrated robust cross-omic performance, achieving balanced accuracy in both transcriptome and proteome datasets. Importantly, substitution of several genes with correlated candidates (*TRMT1L* substituted with *ACOT13, MRPS14* with *NDUFB1*, and *TST* with *MPST*) did not diminish predictive performance, supporting the robustness of the identified biomarker panel.

Our work has several important implications. First, it establishes the feasibility of using proteotranscriptomic integration for radiosensitivity prediction, extending beyond transcriptome-only classifiers such as the gene expression–based radiation sensitivity index (RSI).[30] Second, it identifies candidate biomarkers with dual-level validation (RNA and protein), increasing the likelihood of biological relevance and translational potential. Third, by demonstrating robust model performance across omic layers, this framework provides a template for future pan-cancer analyses and potential clinical applications, where integrating multi-omics assays may enhance predictive precision in radiotherapy.

Nonetheless, some limitations warrant consideration. Our analysis was restricted to in vitro NSCLC cell lines, which do not fully recapitulate the tumor microenvironment, immune interactions, and intratumoral heterogeneity observed in patients. Additionally, the relatively small number of cell lines in the proteomic dataset (n = 46) may limit generalizability, although cross-validation and substitution testing mitigated potential overfitting. Finally, while SVR provided strong predictive performance, future studies should compare alternative machine learning strategies (e.g., ensemble methods, deep learning) and investigate whether integrating additional omics layers (e.g., epigenomics, metabolomics) further improves accuracy.

## 5. Conclusions

In summary, this study demonstrates that concurrent transcriptome–proteome analysis enables the discovery of robust biomarkers and predictive models for NSCLC radiosensitivity. By combining RNA-seq and mass spectrometry–based proteomic data, we identified a robust panel of concurrent biomarkers that capture both transcriptional regulation and functional protein activity. The identified biomarker panel, enriched in genes regulating chromatin remodeling, RNA splicing, and stress response, offers both mechanistic insight and translational promise. Compared with single-omics approaches, the combined model demonstrated superior cross-omic predictive performance. Extending this approach to larger, clinically annotated cohorts and pan-cancer datasets will be a critical next step toward developing precision radiotherapy strategies guided by integrative omics biomarkers.